%% file: 0_main.tex
\documentclass[letterpaper, 10 pt, conference]{ieeeconf}

\IEEEoverridecommandlockouts

\usepackage{etextools}
\usepackage{amssymb}
\usepackage{float}
\usepackage{makeidx}
\usepackage{amsmath}
\usepackage{bbm}
\usepackage{epsfig}
\usepackage{epsf}
\usepackage{psfrag}
\usepackage{verbatim}
\usepackage{color}
\usepackage{multirow}
\usepackage{tabularx}
\usepackage{booktabs}
\usepackage[tight,footnotesize]{subfigure}
\usepackage{array}
\usepackage{soul}
\usepackage{footnote}
\usepackage{cite}
\usepackage{dblfloatfix}
\usepackage[inkscapelatex=false]{svg}
\usepackage{color, colortbl}
\usepackage[colorlinks,bookmarksnumbered,citecolor=orange,urlcolor=orange]{hyperref}
\usepackage{graphicx}
\graphicspath{{Figures}}
\DeclareGraphicsExtensions{.pdf,.png}
\usepackage{bigstrut}
\usepackage[english]{babel}

\usepackage{csquotes}

\usepackage[T1]{fontenc}
\usepackage{algorithm, setspace}
\usepackage{algpseudocode}
\usepackage{url}
\usepackage{multirow}
\usepackage{float}
\usepackage{xcolor}
\usepackage{hyperref}
 \hypersetup{
     colorlinks=true,
     linkcolor=orange,
     filecolor=orange,
     citecolor=orange,      
     urlcolor=orange,
     }

\newcommand{\p}[1]{\smallskip \noindent \textbf{{#1}.}}
\newcommand{\eq}[1]{Equation~(\ref{eq:#1})}
\newcommand{\fig}[1]{Figure~\ref{fig:#1}}
\usepackage{balance}
\usepackage{bbold}

\title{\LARGE
A Pivot-Based Kirigami Utensil for Hand-Held and Robot-Assisted Feeding
}

\author{Keone Leao*, Grace Brotherson*, Iain Mischel*, Sagar Parekh, and Dylan P. Losey
\thanks{*K. Leao, G. Brotherson, and I. Mischel contributed equally to this work.\newline
Collaborative Robotics Lab (\href{https://collab.me.vt.edu/}{Collab}), Dept. of Mechanical Engineering, Virginia Tech, Blacksburg, VA 24061. Email: \texttt{losey@vt.edu}}
}

\begin{document}
\maketitle

\begin{abstract}

Eating is a daily challenge for over $60$ million adults with essential tremors and other mobility limitations.
For these users, traditional utensils like forks or spoons are difficult to manipulate --- resulting in accidental spills and restricting the types of food that can be consumed.
Prior work has developed rigid, hand-held utensils that often fail to secure food, as well as soft, shape-changing utensils made strictly for robot-assisted feeding.
To assist a broader range of users, we introduce a re-designed kiri-spoon that can be leveraged as \textit{either a hand-held utensil or a robot-mounted attachment}.
Our key idea --- developed in collaboration with stakeholders --- is a \textit{pivot-based design}.
With this design the kiri-spoon behaves like a pair of pliers: users squeeze the handles to change the shape of the utensil and enclose food morsels.
In practice, users can apply this kiri-spoon as either a spoon (that scoops food) or as a fork (that pinches food); when the handles are closed, the utensil wraps around the morsel and prevents it from accidentally falling.
We characterize the amount of force required to open or close the kiri-spoon, and show how designers can modify this force through kinematic or material changes.
A highlight of our design is its accessibility: the hand-held version consists of just four $3$D printed parts that snap together.
By adding a servo motor, we can extend this same kinematic structure to robot-assisted feeding.
Across our user studies, adults with disabilities and elderly adults with Parkinson's reported that the kiri-spoon better met their needs and provided a more effective means of spill prevention than existing alternatives.
See a video of our kiri-spoon here: \url{https://youtu.be/FFIomm5RL98}

\end{abstract}


\input{1_intro}

\input{2_related}

\input{3_kinematics}

\input{4_stiffness}

\input{5_study1}

\input{6_study2}

\input{7_conclusion}


\balance
\bibliographystyle{IEEEtran}
\bibliography{IEEEabrv,bibtex}

\end{document}

%% file: 1_intro.tex
\begin{figure*}[t]
	\begin{center}
        \includegraphics[width=2.0\columnwidth]{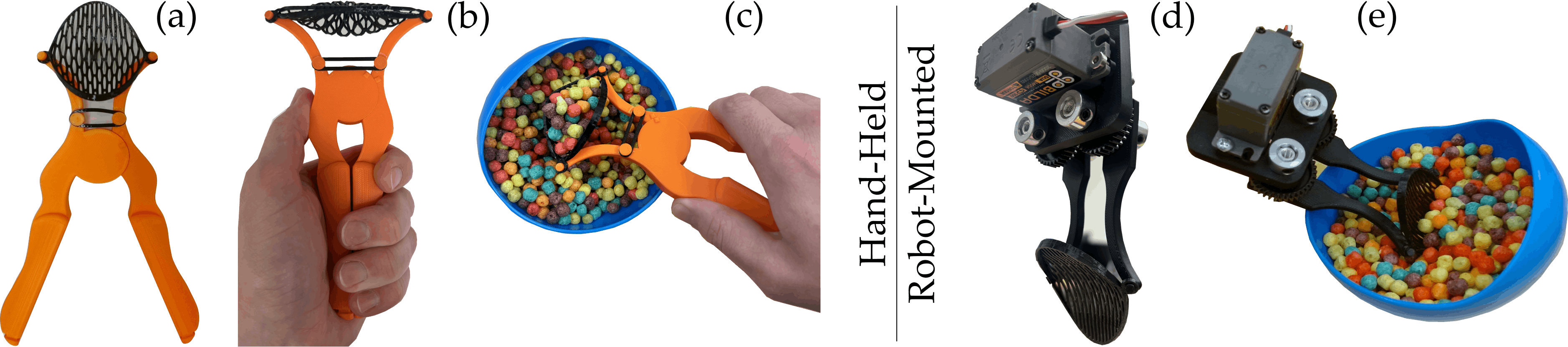}
		\caption{By using a pivot design similar to pliers, we enable hand-held and robot-mounted kiri-spoons. Rotating the pivot causes the kirigami mesh to extend, changing its shape from a flat sheet to an ellipsoid enclosure.  The kirigami enclosure prevents accidental spills. a) Hand-held kiri-spoon. b-c) User squeezing the kiri-spoon to acquire cereal from a bowl. d) Robot-mounted kiri-spoon. e) Robot arm leveraging the kiri-spoon to grasp foods.}
		\label{fig:front}
	\end{center}
    \vspace{-1.5em}
\end{figure*}

\section{Introduction} \label{sec:intro}

How do humans manipulate food?
Traditional utensils such as spoons, forks, or chopsticks enable most adults to seamlessly pick up food items and transfer those morsels to their mouths.
But for the over $60$ million people worldwide with essential tremors, traditional utensils can be difficult to utilize \cite{ETstats}.
If the human's hand has unintentionally jerky motions --- e.g., the effects of Parkinson's, arthritis, or other motor limitations --- they may accidentally spill food from their spoon, or fail to stab a morsel with their fork.

Existing research seeks to address this problem and assist users in two main ways.
First, a variety of \textit{hand-held utensils} have been developed to counteract the effects of motor disabilities.
Modified utensils feature larger grips, angled stems, or gyroscopic stabilization to keep food balanced atop the utensil.
Users can use their own arms to manipulate these custom utensils and eat everyday meals.
A second paradigm is \textit{robot-assisted feeding}, where an external robotic arm picks up the desired food and brings it to the human's mouth.
These robots are usually equipped with traditional utensils, and the robot must coordinate its positions and velocities to manipulate that utensil correctly.

Both paradigms have shown promising results for users with varying levels of motor disability.
However, the mechanical design of today's assistive utensils also presents fundamental limitations.
From the hand-held perspective, modified utensils are either too rigid to prevent food from sliding off, or overly expensive with complex electrical components that require charging \cite{sabari2019adapted}.
From the robotic perspective, it is challenging for robot arms to orchestrate their motions so that they acquire, carry, and transfer morsels without spilling \cite{keely2025kiri}.
Traditional utensils were developed for humans --- not robots --- making them inefficient for robot-assisted feeding.

In this work we introduce a modified spoon that can be used as either a hand-held utensil or robotic attachment.
We begin from the kiri-spoon \cite{keely2025kiri, keely2024kiri}, a robot-mounted device with a soft kirigami structure.
By stretching the kirigami, the kiri-spoon changes shape and encloses food items.
To convert this mechanism into an accessible utensil with both hand-held and robotic versions, our insight is:
\begin{center}
    \textit{We can articulate the utensil using a pivot-based design where users squeeze the handle to stretch the kirigami.}
\end{center}
Applying this design principle we present a versatile tool that can be used either as a spoon (to scoop food) or as a fork (to pinch morsels).
The hand-held version is fully $3$D printed, and consists of just four parts (see \fig{front}, left).
We characterize the amount of force required to close the device, and co-design with stakeholders to ensure the tunable squeezing motion is suitable.
This hand-held utensil is different from today's assistive alternatives because it (a) has a soft, shape-changing structure and (b) when squeezed, it holds food within the kirigami mesh, preventing them from falling despite user tremors or spasticity.  

The same kinematic and material structure also extends to robot-assisted feeding (see \fig{front}, right).
Here we add a single servo motor to open or close the kirigami mesh around the pivot point, and attach the resulting mechatronic device to the robot's end-effector.
Compared to previous kiri-spoons, our new approach to actuation greatly simplifies the system and introduces a unified, self-contained mechatronic design that can be reproduced by other researchers.
Depending on the stakeholder's specific motor disabilities, either the hand-held or robotic version may be more suitable.

Overall, we see this work as a step towards accessible utensils for assistive eating, whether manipulated by humans or robot arms.
We make the following contributions:

\p{Hand-Held Design}
We present the kinematics and manufacturing process for our hand-held kiri-spoon.
This design consists of four components that can be $3$D printed with PLA and TPU and snapped together.
We include the files needed to print these kiri-spoons, and also derive the relationship between applied forces and kirigami displacement.

\p{Mechanics Characterization}
We model the kirigami sheet as a spring, and estimate its spring constant.
We find that the force required to stretch the sheet primarily depends on the mesh material, and support our theoretical model with simulations and real-world experiments.

\p{Multiple User Studies}
We conduct three separate evaluations.
These experiments span stakeholders living with long-term disability, elderly adults with Parkinson's, and adults without disabilities.
From our user studies we identify key design parameters, and compare our utensil to other customized alternatives.
Overall, the results suggest that the kiri-spoon (whether hand-held or mounted to a robot) is more effective at preventing spills while remaining user-friendly.

%% file: 2_related.tex
\section{Related Works} \label{sec:related}

We focus on assistive feeding utensils.
Existing approaches can be sorted into two categories for stakeholders with varying needs: hand-held utensils and robot-assistive feeding.

\p{Modified Hand-Held Utensils}
There are commercial utensils for users with tremors or irregular motor control \cite{rehabmarket, pagnussat2025people}.
The two best performing types are weighted utensils, which passively dampen high-frequency tremors, or self-stabilizing utensils, which use onboard electronics to align the handle and tip \cite{sabari2019adapted}.
For example, the Liftware system \cite{liftware, eli} tries to prevent food from spilling by re-balancing the utensil so that it always remains horizontal.
However, there are several shortcomings with these approaches.
Adding weight only helps a specific set of users \cite{sabari2019adapted}, and does nothing to prevent food from accidentally sliding off a rigid spoon.
Liftware and its alternatives cost upwards of $200$ USD, require daily charging, rely on manual switches between spoon and fork heads, and can prevent the user from tilting their utensil --- even when this tilt is intended \cite{liftware, eli}.
Unlike these \textit{rigid} hand-held approaches, we propose a \textit{soft, shape-changing} alternative that wraps around target morsels, preventing falls regardless of the utensil weight or handle orientation.

\p{Robot-Assisted Feeding}
For users who require external assistance when eating, robot arms can automate the process of acquiring bites of food and carrying them to the user's mouth \cite{liu2025robot, nanavati2025lessons}.
Commercial systems have been developed specifically for this purpose \cite{obi, neater}, and ongoing research enables multi-purpose robot arms to also provide feeding assistance \cite{tai2025grits, padmanabha2025waffle, tsakona2025continuous, gordon2023towards}.
Here we focus on the \textit{utensils} that these robots leverage.
In most prior works the assistive robot uses a traditional fork or spoon \cite{park2020active, yow2025franc}, sometimes incorporating extra degrees-of-freedom to increase the utensil's workspace \cite{jenamani2024flair, sundaresan2023learning}.
But there has been a recent push towards customized utensils for robot-assisted feeding.
The advantage of these modified utensils is that they make it easier for the robot arm to acquire and carry food, while still maintaining a user-friendly form factor.
Examples include an origami-based spoon \cite{song2025soda} and utensils that break during failures \cite{chang2025design}.
Most relevant to our work is the original kiri-spoon developed in \cite{keely2024kiri, keely2025kiri}.
Our approach builds on these prior \textit{robot-mounted} utensils to create a \textit{unified, pivot-based design}.
This yields both an accessible hand-held version (that requires no electronics and can be fully $3$D printed) and a simplified mechatronic version for robot-assisted feeding.

%% file: 3_kinematics.tex
\section{Kinematics and Design} \label{sec:kinematics}

Prior work developed a robot-mounted version of the kiri-spoon \cite{keely2025kiri}.
In this early version the kiri-spoon was actuated by a motor which retracts one end of the kirigami sheet; the other end is held in place by a metal wire that wraps around the device.
There are two key issues with this kiri-spoon: a) the design requires onboard and offboard electronics, making it unsuitable for hand-held use, and b) the user must eat around the metal wire, leading to uncomfortable bites.
In this section we present a design which addresses both challenges, enabling hand-held or robot-mounted kiri-spoons.

\subsection{Hand-Held Version} \label{sec:K1}

We begin by focusing on the hand-held kiri-spoon; this design is intended for users who can eat on their own, but struggle to manipulate traditional utensils.
Our finalized design is shown in \fig{front}.
The hand-held version does not require electronics, is inexpensive and food-safe, and can be cleaned by a dishwasher.

\p{Actuation Mechanism}
The proposed kiri-spoon is based on a simple pivot that the user squeezes.
The system functions like pliers: when the user applies forces to both handles, the opposite sides are pulled apart, causing the kirigami mesh to extend.
Because of our cut pattern within the kirigami mesh, extension leads to a change in shape: the spoon increases curvature and wraps around food items.

Our pivot-based actuation mechanism addresses the key problems with \cite{keely2025kiri}.
It greatly simplifies the design, reduces the number of components, and removes the need for any metal wires.
Now the only part of the kiri-spoon that enters the user's mouth is the kirigami mesh; we note that this mesh is deformable, and users can manipulate it with their mouths as necessary.
The intended procedure is for the user to reach for food with the kiri-spoon open, and then squeeze the kiri-spoon to \textit{capture} morsels.
If the user holds the handle closed while carrying food to their mouth, the kirigami mesh prevents any morsels from sliding out and accidentally spilling\footnote{The kirigami mesh can be printed with a porous or continuous base. Our experiments show that a porous base is sufficient for most food, but a continuous mesh is necessary for liquids like soup. See \fig{mesh}.}.
The user then opens the kiri-spoon to transfer the bite into their mouth.
In practice, users can leverage the kiri-spoon as either a spoon (scooping food) or as a fork (pressing down on food).
The key limitation is that the desired morsel must be smaller than the kirigami enclosure.
However, other hand-held assistive spoons face the same limitation, and food for adults with motor impairments is usually cut into bite-sized portions before serving \cite{bhattacharjee2020more}.

\p{Stakeholder Feedback}
The design we present was created by incorporating feedback from five ($N=5$) stakeholders living in The Virginia Home \cite{VirginiaHomePurpose}, a residential community for adults with physical disabilities.
These residents tested different variants of the hand-held kiri-spoon, and provided informal suggestions and guidance on necessary improvements.
Much of their feedback focused on the act of opening or closing the utensil.
Interestingly, stakeholders found it easy to close the kiri-spoon, but \textit{struggled to release their grip so that the spoon could re-open}.
Based on their guidance, we added a 3D printed elastic band to the hand-held kiri-spoon.
This band behaves like a spring to make the open state an equilibrium point: users must apply more force to squeeze the kiri-spoon and enclose food, but by reducing their applied force the spoon naturally re-opens.
While this design requires sustained grip force when carrying food, stakeholders explained that maintaining a closed grip was significantly easier for them than opening their hands or, for example, executing the fine-motor control required to trigger a mechanical latch.
The band can be added or removed based on user need, and designers can also adjust the stiffness of this band to better suit an individual stakeholder.

\p{Manufacturing and Assembly}
The hand-held design consists of four $3$D printed parts: two handles, the kirigami mesh, and the band.
In our experiments the handles were printed using polylactic acid filament (PLA), and the kirigami mesh and band were printed using thermoplastic polyurethane (TPU).
Both PLA and TPU are non-toxic, food-safe, and common materials for standard $3$D printers \cite{PLAinfo_ProtolabsNetwork, TPUinfo_ProtolabsNetwork}.
Using PLA makes the handles rigid, while using TPU makes the kirigami mesh and band elastic.
This elasticity is repeatable: because of the high tensile strength of TPU, we were able to cycle a single kiri-spoon across multiple sessions without any noticeable degradation in the elastic components.
Assuming the user has access to a $3$D printer, the hand-held kiri-spoon can be manufactured using $43$g of filament for a total cost of $\approx 1.57$ USD.
We include part files and detailed assembly guides here:
\url{https://github.com/Kiri-Spoon/Kiri-Spoon.git}

\begin{figure}[t]
	\begin{center}
        \includegraphics[width=1.0\columnwidth]{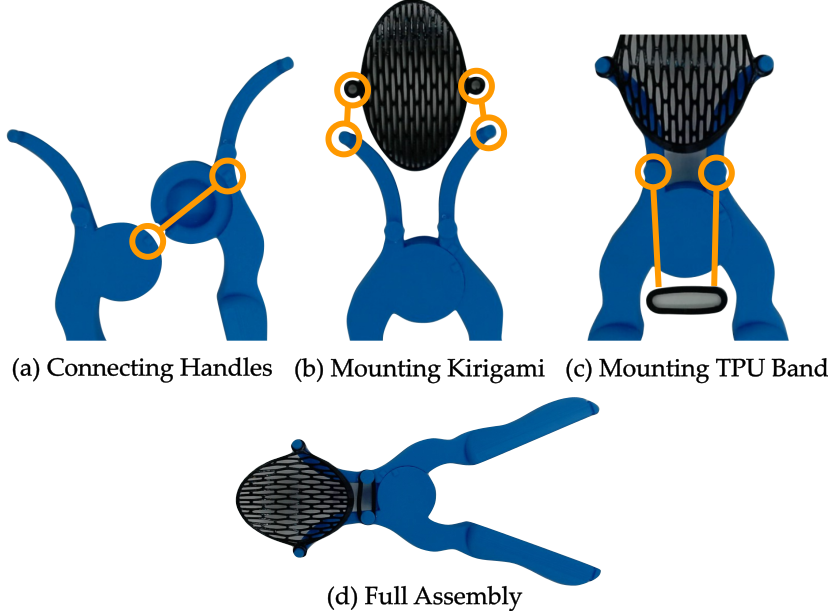}
        \caption{Assembling a hand-held kiri-spoon. (a) The handles can be snapped together by aligning the two bumps. (b) The kirigami mesh is then attached by pushing its side holes onto their respective pegs. (c) A band is wrapped around the posts. This optional band acts as a spring to open the mesh when the squeezing force is removed. (d) The completed assembly.}
		\label{fig:assembly}
	\end{center}
    \vspace{-1em}
\end{figure}

We summarize the assembly process in \fig{assembly}.
Users first snap together the two handles to form the central pivot point --- we have added small bumps to the surfaces of those handles to help guide this process.
Once the handles are attached, the user then presses the kirigami mesh onto the pegs located at the end of the handles.
This kirigami mesh has a gradient design, where the back is thicker than the front to better encapsulate food.
Finally, the user wraps the TPU band around the marked posts.
The kiri-spoon is ready to use in this form; if preferred, the user can also print out and slide on a strap that goes around their hand and prevents the human from accidentally dropping the entire utensil.

\p{Force vs. Displacement}
Now that we have a design for the hand-held kiri-spoon, we next want to evaluate the kinematics and physics of this design.
Specifically, we are interested in the relationship between the user's applied force and the extension of the kirigami mesh.
Put simply: how much force does the user need to apply to close the spoon and encapsulate food?

\begin{figure}[t]
    \begin{center}
        \includegraphics[width=0.5\textwidth]{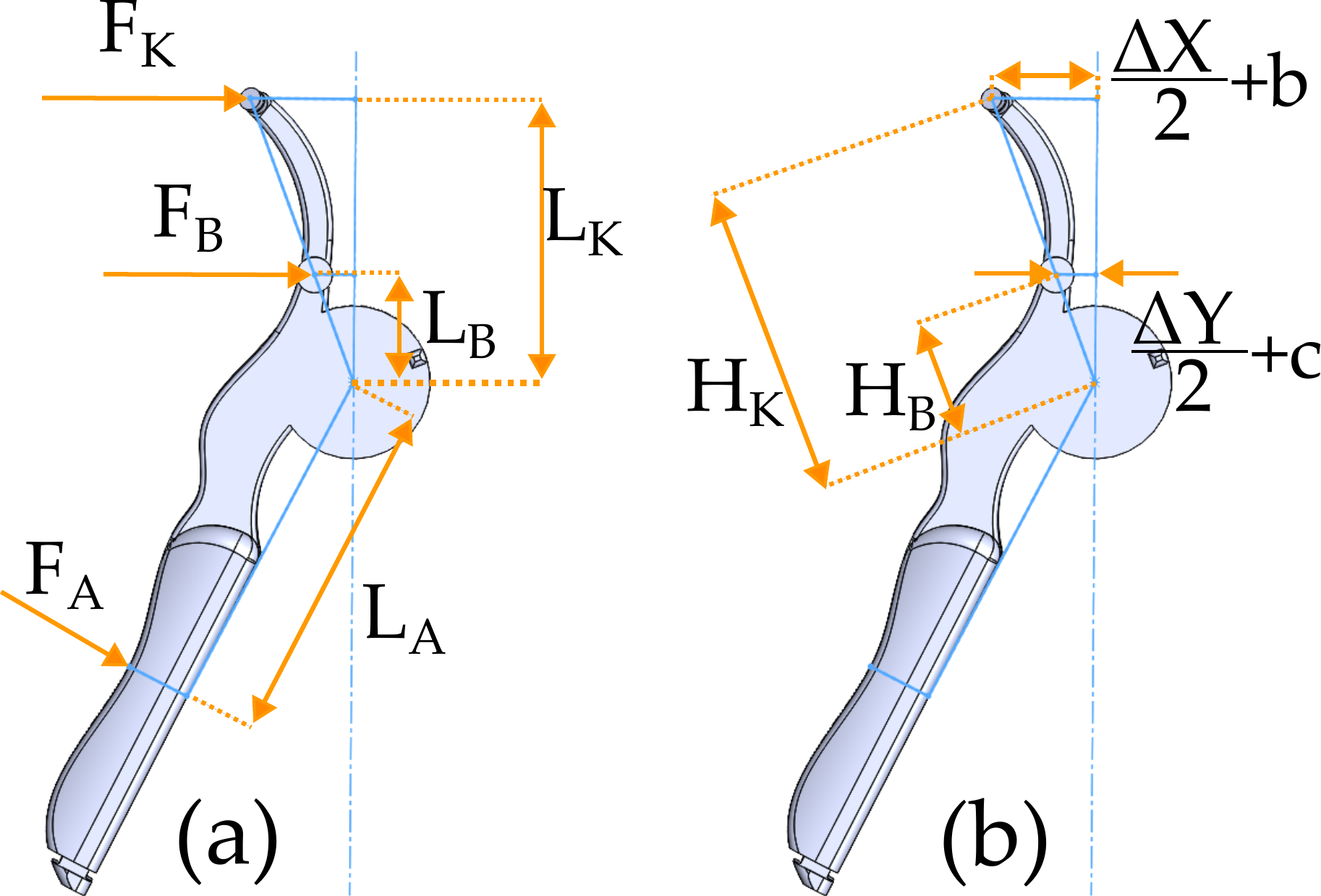}
        \caption{Schematic of the applied forces, moments, and generated displacements. (a) The forces of the kirigami and band are applied horizontally to their respective pegs, while the squeezing force is applied perpendicular to the handle. (b) The blue lines represent the moment arms and similar triangles utilized in \eq{K2}. The central vertical line represents the line of symmetry for the full kiri-spoon assembly.}
        \label{fig:kinematics}
    \end{center}
\end{figure}

Since the assembly is symmetrical, we start by measuring the forces applied to a single handle.
As shown in \fig{kinematics}(a), the three forces acting on the kiri-spoon’s handle are the human's applied force ($F_A$), the kirigami mesh force ($F_K$), and TPU band force ($F_B$). 
For our analysis we will assume that $F_A$ is applied perpendicular to the thinnest part of the handle.
When multiplied by their respective moment arms ($L_A$, $L_K$, $L_B$), each force produces a moment about the pivot point.
At static equilibrium, the human's applied moment is equal to the kirigami and band moments:
\begin{equation} \label{eq:K1}
    F_A L_A = F_K L_K + F_B L_B
\end{equation}
The TPU band can be modeled as a pure spring with stiffness $K_B$.
For example, in our experiments we test a band with stiffness $2.18$ N/mm.
As we will show in Section~\ref{sec:mechanics}, we can also model the kirigami mesh as a spring.
Here the spring constant is $K_K E$, where $K_K$ is the kirigami stiffness factor and $E$ is Young’s Modulus for the kirigami material.

\begin{table}[t]
    \centering
    \begin{tabular}{c|c|c}
       \textbf{Variable} & \textbf{Description} &  \textbf{Value}\\
       $L_A$ &Applied Force Moment Arm &$69.6$ mm  \\
       $K_K$ &Kirigami Stiffness Factor & $4.55$ mm  \\
       $E$ & Kirigami Young's Modulus & $14.9$ MPa  \\
       $H_K$& Kirigami Hypotenuse  & $59.2$ mm  \\
       $b$ & Kirigami Offset &$20.7$ mm  \\
       $K_B$ & Band Stiffness &$2.18$ N/mm  \\
       $H_B$ & Band Hypotenuse &$22.5$ mm  \\
       $c $ & Band Offset & $7.8$ mm  \\  
    \end{tabular}
    \caption{Constants for the Force-Displacement Equation.}
    \label{table:variables}
    \vspace{-1em}
\end{table}

We will therefore use Hooke's law to model $F_K$ and $F_B$.
What makes this challenging is that the moment arms for both the kirigami mesh ($L_K$) and the band ($L_B$) change as the kiri-spoon is actuated.
Let $\Delta X$ be the horizontal displacement of the kirigami mesh: this is a key variable, since it measures how much the human has opened or closed the spoon.
Similarly, we define $\Delta Y$ as the horizontal displacement of the TPU band.
Leveraging the geometry in \fig{kinematics}(b), we can write the changing moment arms in terms of the spring displacements $\Delta X$ and $\Delta Y$:
\begin{multline} \label{eq:K2}
    F_A L_A = K_K E \Delta X\sqrt{H_K^2-(\Delta X+b)^2}
    \\ + K_B\Delta Y \sqrt{H_B^2-(\Delta Y+c)^2}
\end{multline}
In \eq{K2}, $H_K$ is the hypotenuse of the triangle defined by the kirigami mesh and $H_B$ is the hypotenuse of the triangle defined by the band.
The constant $b$ is the initial offset between the kirigami attachment point and the neutral axis; similarly, $c$ is the offset between the band attachment point and the neutral axis.
We provide a list of these variables and their values
for an example kiri-spoon in Table~\ref{table:variables}.

Our goal is to derive a relationship between the human's applied force, $F_A$, and the kirigami displacement, $\Delta X$.
To complete this relationship we use the property of similar triangles.
Recognizing that $\Delta Y + c = (H_B / H_K) (\Delta X + b)$, we obtain our final result:
\begin{multline*}
    F_A L_A = K_K E \Delta X\sqrt{H_K^2-(\Delta X+b)^2}
    \\ + K_B\Bigg(\frac{H_B}{H_K}(\Delta X + b) - c\Bigg) \sqrt{H_B^2-\bigg(\frac{H_B}{H_K}(\Delta X+b)\bigg)^2}
\end{multline*}
The first term on the right side of this equation captures the forces applied by the kirigami mesh, and the second term captures the forces applied by the TPU band.
If the band is removed, then the force-displacement relationship simplifies to: $F_A L_A = K_K E \Delta X\sqrt{H_K^2-(\Delta X+b)^2}$.
Designers can leverage this equation to tune the overall stiffness of their own hand-held kiri-spoon.
By scaling the overall size up or down, designers change the hypotenuses and offsets (altering $H_B$, $H_K$, $b$, $c$).
Alternatively, the designer can adjust the stiffness by changing the materials used to print the kirigami mesh and band (altering $E$ or $K_B$).

\subsection{Robot-Mounted Version} \label{sec:K2}

In Section~\ref{sec:K1} we presented our design for a hand-held kiri-spoon. 
Here we extend these same design principles to a robot-mounted version.
Most of the design details are identical across both versions; the primary difference is the actuation.
For the robot-mounted kiri-spoon we use a servo motor to turn the pivot and open or close the kirigami mesh.
As shown in \fig{front} (right), this kiri-spoon can then be mounted to the end-effector of a robot arm.
All the parts and assembly instructions are available in our repository; the system can be manufactured for $\approx 102.55$ USD (both the motor and the gears are around $35$ USD each).

%% file: 4_stiffness.tex
\begin{figure*}[t]
	\begin{center}
        \includegraphics[width=2.0\columnwidth]{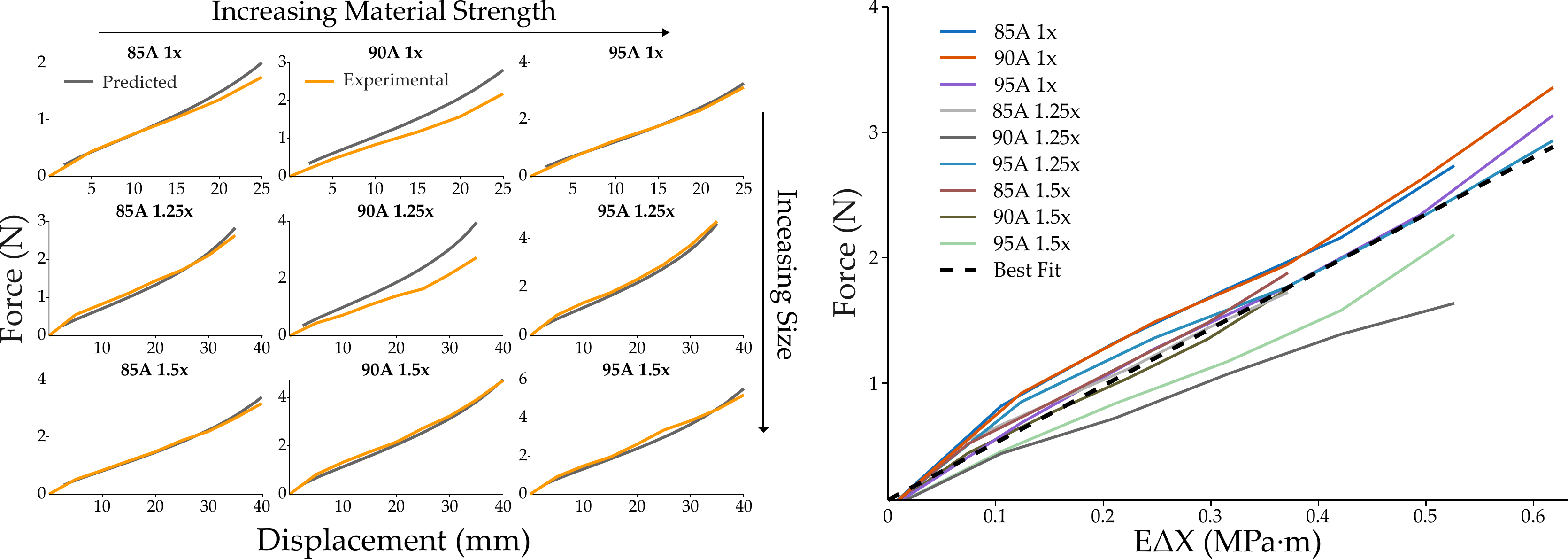}
		\caption{Results from Section~\ref{sec:mechanics} used to model the kirigami mesh. (Left) Force vs. Displacement from real-world experiments and ANSYS simulations. Each separate graph represents a change in the kirigami sheet's parameters. From top to bottom, the sheet is increasing in size from $1\times$ to $1.5\times$. From left to right, the sheet is increasing in material stiffness, ranging from a Shore hardness of $85$A to $95$A. (Right) Force vs. Young's modulus times displacement. Here we group all nine experimental plots on a single graph, and show that (a) we can approximate the kirigami mesh as a spring, and (b) the spring constant is of the form $K_K E$. A line of best fit is overlaid to highlight the kirigami stiffness factor $K_K$ from our model.}
		\label{fig:stiffness}
	\end{center}
    \vspace{-1.5em}
\end{figure*}

\section{Modeling the Kirigami Mesh} \label{sec:mechanics}

The core of our spoon is the kirigami mesh.
By squeezing the handles --- or turning the pivot --- we can extend both sides of the kirigami, causing it to deform.
Looking at \fig{front}, the kirigami has a complex pattern of interconnected bands and ribbons.
These ribbons were designed so that the kirigami increases curvature when extended, ultimately going from a flat sheet to a container for the food items.
Despite its complexity, \textit{we hypothesize that we can model the overall kirigami mesh as a spring}.
More specifically, here we show that the spring constant only depends on the material properties, and not on the size of the kiri-spoon.

\p{Independent Variables}
For testing the kirigami mesh we varied two parameters.
First, we changed the \textit{materials} used to $3$D print the mesh: we manufactured kirigami sheets using TPU with Shore hardness $85$A, $90$A, and $95$A.
We then performed tensile tests to measure the Young's modulus ($E$) for each of these materials.
The second parameter we varied was the \textit{size} of the kirigami mesh.
We scaled up our geometry from $1\times$, to $1.25\times$, and finally to $1.5\times$.
These two parameters are important because users will manufacture their kiri-spoons out of different materials, and may change the size of their kiri-spoon to suit their individual needs.
We therefore wanted to determine how $E$ and the mesh size affect the kirigami sheet's overall stiffness.

\p{Testing Procedure}
Our testing procedure measured the force required to stretch the kirigami mesh.
For each combination of mesh material and size, we increased the displacement $\Delta X$, and recorded the force $F_K$ required to maintain that displacement.
We only tested displacements that were within our operating range (i.e., displacements between a flat kirigami sheet and its enclosed state).
Displacements beyond this limit are highly non-linear, but are not relevant for users who just want to open or close the kiri-spoon.
Our tests were performed using both physical components and finite-element simulations: the results are summarized in \fig{stiffness}.

To collect the \textit{Experimental} curve we mounted one end of the kirigami mesh to a fixed point and attached a force sensor to the opposite side.
We then stretched the mesh for displacements ranging from $5$ mm to $50$ mm, with $5$ mm increments, and recorded the force at each displacement.
This testing was repeated three times to obtain average values.

To collect the \textit{Theoretical} curve we used ANSYS simulations.
We first built an accurate version of our kirigami sheet in ANSYS by using the neo-Hookean hyper-elastic model \cite{martin2025anisotropic}.
We then obtained the shear modulus and bulk modulus from our measured $E$ values, and set $0.003$ MPa$^{-1}$ as our incompressibility factor.
Finally, we actuated this model to obtain the forces $F_K$ at different displacements $\Delta X$.

\p{Spring Model}
Our individual results for each material and size are displayed on the left of \fig{stiffness}.
These plots show a strong agreement between experimental and theoretical stiffness, and the highly linear force-displacement relationship suggests that the mesh can be modeled as a spring.
To determine what factors affect that spring constant, we turn to the right side of \fig{stiffness}.
Here we show all nine of the experimental results on a single plot, where the $x$-axis is $E\Delta X$, and the $y$-axis is the resulting force.
Based on this plot we argue that \textit{the size of the mesh does not noticeably impact kirigami stiffness}.
We recognize that size may have an impact if it is significantly changed --- but for our utensil application, the size is restricted to standard spoons.
Instead, the stiffness of the kirigami sheet is based on material:
\begin{equation} \label{eq:M1}
    F_K = K_K E \Delta X
\end{equation}
A value of $K_K = 4.55$ mm was estimated from our experimental data using linear regression.
Note that the units of $K_K$ make sense when multiplied by Young's modulus --- i.e., $K_K \cdot E$ has units N/mm.

\p{Impact for Designers}
Based on our experimental and theoretical results we approximate the kirigami mesh as a spring.
\eq{M1} provides designers with insight on how to modify the stiffness of this spring --- specifically, designers can change the kirigami stiffness by altering the material.
Plugging \eq{M1} back into the kinematics analysis from Section~\ref{sec:K1} provides an overall understanding of how much force humans must apply to actuate the hand-held kiri-spoon.
We recommend that designers identify their individual user needs (e.g., does the user struggle to open or close the kiri-spoon?), and then modify the mesh material $E$ and/or band stiffness $K_B$ to suit that user.

%% file: 5_study1.tex
\section{Preliminary Study: \\ University Students without Disability} \label{sec:user1}

\begin{figure*}[t]
	\begin{center}
        \includegraphics[width=2.0\columnwidth]{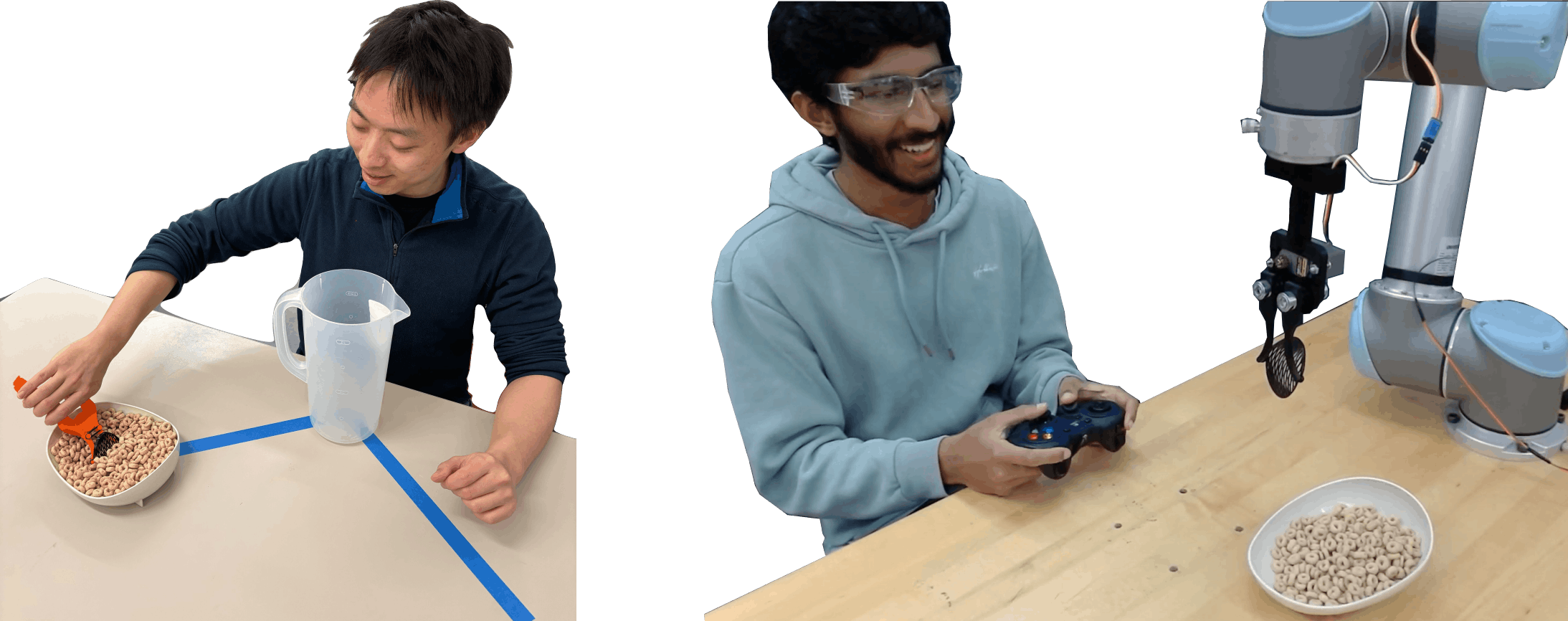}
		\caption{Setup for our preliminary study with university students without mobility limitations. (Left) In the \textbf{Manual} phase participants completed two tasks using hand-held utensils: transferring food from a bowl to a pitcher, and eating food from the bowl. (Right) In the \textbf{Robotic} phase we mounted the utensil to the end-effector of a UR5 robot arm. Using a combination of teleoperation and programmed motions, participants controlled the robot to eat cereal from a bowl. We compared our pivot-based kiri-spoon to a Liftware utensil in both phases.}
		\label{fig:userstudy}
	\end{center}
 	\vspace{-1.5em}
\end{figure*}

We have presented our pivot-based re-design of the kiri-spoon, and characterized its force-displacement relationship.
The objective of this modified utensil is to enable individuals with motor limitations to manipulate food without spilling.
To evaluate the feasibility of our design, we begin with a proof-of-concept study with participants without disability.
The study compares both hand-held and robot-mounted versions of our kiri-spoon against Liftware \cite{liftware}, a self-balancing utensil currently available for purchase.

\p{Experimental Setup and Procedure} 
The experiment was divided into two phases corresponding to the two applications of the kiri-spoon: hand-held (\textbf{Manual}) and robot-mounted (\textbf{Robotic}).
In both \textbf{Manual} and \textbf{Robotic} participants scooped food from a bowl containing $50$ g of Honey Nut Cheerios.
We compared two conditions: using the Liftware spoon and using the kiri-spoon.
To avoid potential ordering effects and learning bias, the order in which the two spoons were evaluated was counter-balanced across participants.

The \textbf{Manual} phase is shown in \fig{userstudy} (left).
Participants held the feeding instrument with their hand to (a) transfer food from a bowl to a pitcher and (b) scoop food from a bowl to feed themselves.
Participants were seated at a table with the pitcher directly in front of them --- the bowl containing the cereal was placed diagonally to the left or right depending on participant's handedness.
Prior to the recorded trials participants were given $5$ minutes to familiarize themselves with the utensils.
In the first task, participants transferred cereal from the bowl to the pitcher over $5$ repetitions.
Next, we replaced the pitcher with a refilled bowl, and instructed participants to eat $5$ scoops of cereal.

The \textbf{Robotic} phase is shown in \fig{userstudy} (right).
In this phase the feeding utensil was attached to the end-effector of a $6$ degree-of-freedom UR5 robot manipulator.
For the Liftware condition, we mounted only the passive spoon attachment to the robot to prevent interference with the Liftware's internal stabilization.
Participants teleoperated the robot to scoop cereal from a bowl.
This teleoperation was performed with a hand-held joystick; when using the kiri-spoon, participants pressed a trigger on the joystick to actuate the servo motor and open or close the kiri-spoon.
Once the user acquired their food, they pressed another button which automated the process of moving the utensil to their mouth.
Users could then use the joystick to adjust the robot's final position for more natural bites.
We provided participants with a practice session to familiarize themselves with the robot and joystick-based teleoperation.
Participants then completed $5$ trials of robot-assisted feeding.

\p{Dependent Variables} 
To assess the effectiveness of each utensil we measured the time taken to acquire and transfer the cereal (\textit{Time}) and the amount of the cereal that was spilled (\textit{Spill}).
\textit{Time} was measured as the total duration from the initiation of the first scoop to the completion of the final scoop.
We computed \textit{Spill} by counting the total number of Cheerios that were dropped over all $5$ repetitions of the task.
Ideally, participants are able to minimize the \textit{Time} they need to manipulate food while also minimizing the amount of food they \textit{Spill} with the utensil.

\p{Participants}
We recruited $12$ participants without disability from the university community.
All participants were undergraduate or graduate students ($4$ female, $8$ male, average age $22.3 \pm 2.9$).
Each participant provided informed written consent according to the university guidelines (IRB $\#$ANON).
We followed a within-subject design where each participant performed \textbf{Manual} and \textbf{Robotic} experiments with both feeding utensils (kiri-spoon and Liftware).

\begin{figure}[t]
	\begin{center}
        \includegraphics[width=1.0\columnwidth]{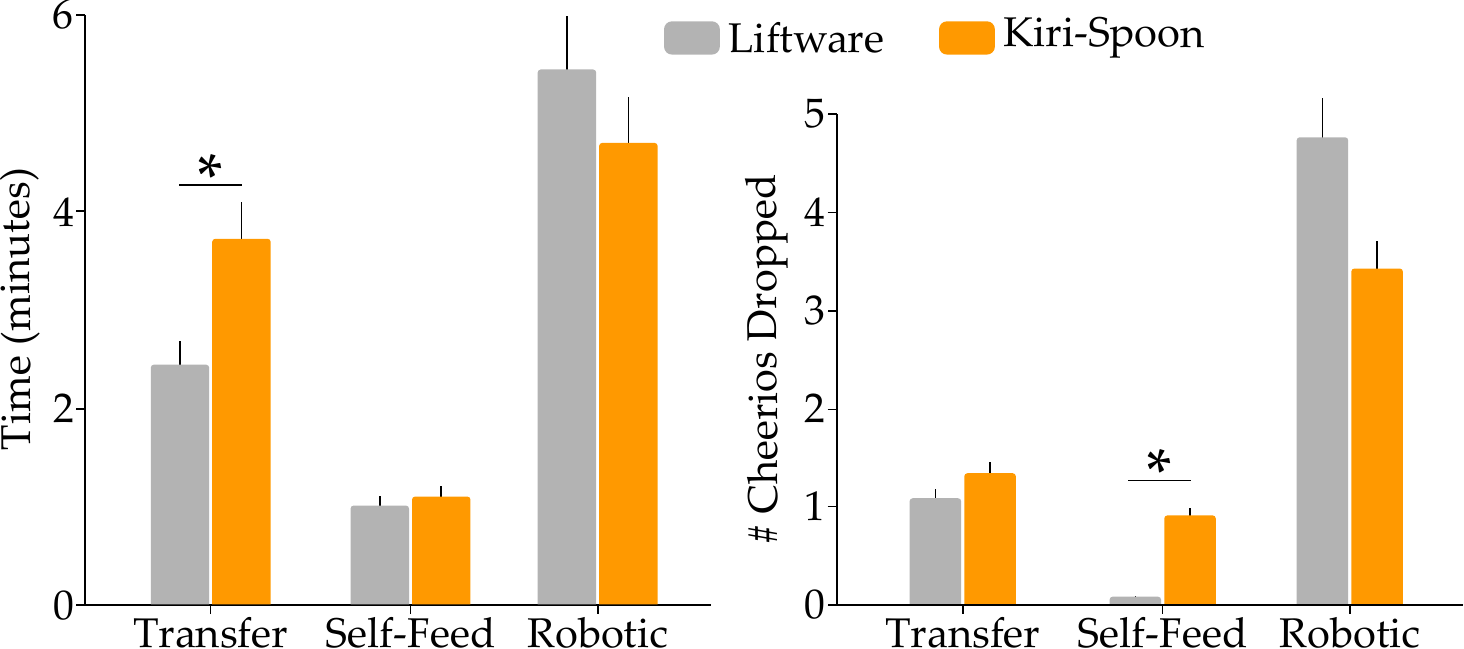}
		\caption{Results from our preliminary study in Section~\ref{sec:user1}. This experiment was performed by users without disability. There were two phases as shown in \fig{userstudy}: \textbf{Manual} and \textbf{Robotic}. We quantify performance using the total \textit{Time} to finish $5$ repetitions of each task --- transferring cereal from a bowl to a pitcher, eating cereal with a hand-held utensil, and eating cereal with a robot-mounted utensil. We also report the average number of Cheerios \textit{Spilled} during each of the tasks. For both metrics we find that the \textbf{Manual} results are better for Liftware, and the \textbf{Robotic} results are better for kiri-spoon. Since the Liftware spoon functions similarly to a standard spoon --- and our participants here had no mobility limitations --- it is not surprising that they were more effective with the hand-held Liftware as compared to the hand-held kiri-spoon. The asterisk (*) indicates statistical significance.}
		\label{fig:results}
	\end{center}
 	\vspace{-2.0em}
\end{figure}

\p{Results}
Our results are summarized in \fig{results}.
After participants completed each phase, we weighed the amount of food for consistency.
With Liftware participants transferred slightly more food ($\approx 1.95$ g total), but an independent samples t-test revealed that the differences were not statistically significant ($p = .17)$.
This suggests participants were carrying similar amounts of food with both utensils.

The \textit{Time} taken to complete the task varied between \textbf{Manual} and \textbf{Robotic} conditions.
In the \textbf{Manual} tasks people were faster with the Liftware utensil: $2.43$ minutes (transfer) and $1$ minute (feeding) with Liftware vs. $3.71$ minutes (transfer) and $1.1$ minutes (feeding) with kiri-spoon.
This trend reversed in the \textbf{Robotic} phase: participants took $5.45$ minutes with Liftware, but only $4.7$ minutes with kiri-spoon.

We observed a similar pattern for the number of \textit{Spills}.
In \textbf{Manual} tasks the participants spilled fewer Cheerios when using the Liftware utensil.
This difference was slight during the transfer task ($1.08$ Cheerios on average with Liftware vs. $1.33$ with kiri-spoon), but the difference became more pronounced during feeding ($0.08$ with Liftware vs. $0.91$ with kiri-spoon).
On the other hand, when performing the \textbf{Robotic} trials users spilled more food with the Liftware utensil.
Here participants spilled $4.75$ Cheerios on average with Liftware, and only $1.9$ with kiri-spoon.

\p{Summary}
These results should be taken with a grain of salt.
The kiri-spoon was developed for users with disabilities, and the tests summarized above were conducted with able-bodied participants.
In practice, the Liftware utensil behaves very similarly to a standard rigid spoon --- it is therefore not surprising that our users found it easier to manipulate the Liftware utensil during \textbf{Manual} tasks.
Put another way: we should not expect our modified utensil to outperform a normal spoon, especially not when the participants have used spoons their entire lives.
Where the kiri-spoon did excel, however, was the \textbf{Robotic} condition: here the kiri-spoon reduced both \textit{Time} and \textit{Spills}.
This result suggests that robot-assisted feeding can be more effective with kiri-spoons, even for users without disabilities.

%% file: 6_study2.tex
\section{Investigative Survey: \\ Elderly Adults with Parkinson's} \label{sec:user2}

In Section~\ref{sec:kinematics} we co-designed with stakeholders, and in Section~\ref{sec:user1} we performed controlled tests with adults without disability.
Now we return to our target population to test our finalized kiri-spoon.
Specifically, we conduct an investigative survey in collaboration with \href{https://retire.org/about/}{Warm Hearth}, a retirement village offering support groups for individuals with disabilities.
Due to safety protocols and logistical constraints, this round of stakeholder evaluation was limited to the \textit{hand-held kiri-spoon}; future work will evaluate the robot-mounted system with the same demographic.

We collaborated with three ($N=3$) elderly residents diagnosed with Parkinson's disease.
These participants provided informed written consent following IRB $\#$ANON.
The tremors and impaired motor control caused by Parkinson's can make eating challenging: two of our volunteers reported occasional difficulty while eating, and the remaining participant reported that they had difficulty eating on a daily basis.
We first introduced the participants to the kiri-spoon through a detailed explanation of its mechanism, the assembly process, and its intended use.
We next provided all three participants with hand-held kiri-spoons for open-ended testing.
Each participant was given $\approx 15$ minutes to physically examine the design, assess the ergonomic comfort, and test it by consuming different food items.
This \textit{user-guided interaction} allowed the participants to provide informed and experience-based feedback.

\p{Survey Results}
After the user-guided interaction we asked participants to rate three statements on a $1$-$7$ scale.
Lower numbers indicated that the participant \textit{disagreed} with the statement, while higher numbers indicated stronger \textit{agreement}.
The statements and the responses are listed below:
\begin{itemize}
    \item ``This spoon was comfortable.'' (scores $7$, $7$, $5$)
    \item ``I found this spoon easy to use.'' (scores $6$, $6$, $6$)
    \item ``This spoon dropped food frequently.'' (scores $1$, $2$)
\end{itemize}
Note that the last statement only had two responses because one participant wanted to test a wider variety of foods before committing to an answer.

Alongside the structured survey we also encouraged the participants to provide free-form feedback.
The participants explained that the kiri-spoon's design ``\textit{held a good amount of food}'' while effectively restricting the portion size (i.e., the amount of food per bite was limited by the volume of the actuated kirigami mesh).
The participants emphasized that this restriction is advantageous because appropriate portion sizes can reduce choking hazards.
One volunteer stated that the spoon was ``\textit{very comfortable},'' and that they ``\textit{would take the kiri-spoon to a restaurant}.''
Participants did not report fatigue or difficulty opening or closing the kiri-spoon.

Our discussion with the participants also highlighted the need for \textit{personalized} utensils.
One user wanted the kiri-spoon to be narrower, while others thought the current shape was suitable.
We recognize that one size will not fit all, and recommend that users scale up or down the provided kiri-spoon files to suit their individual needs.
Overall, the feedback we obtained from elderly adults with Parkinson's indicated that (a) the kiri-spoon was comfortable and easy to use, (b) the kiri-spoon prevented accidental spills, and (c) individual users will want to tune their own kiri-spoon parameters.
Despite the small sample size, these findings support the key contributions of our hand-held design.

%% file: 7_conclusion.tex
\section{Conclusion} \label{sec:conclusion}

\begin{figure}[t]
	\begin{center}
		\includegraphics[width=1.0\columnwidth]{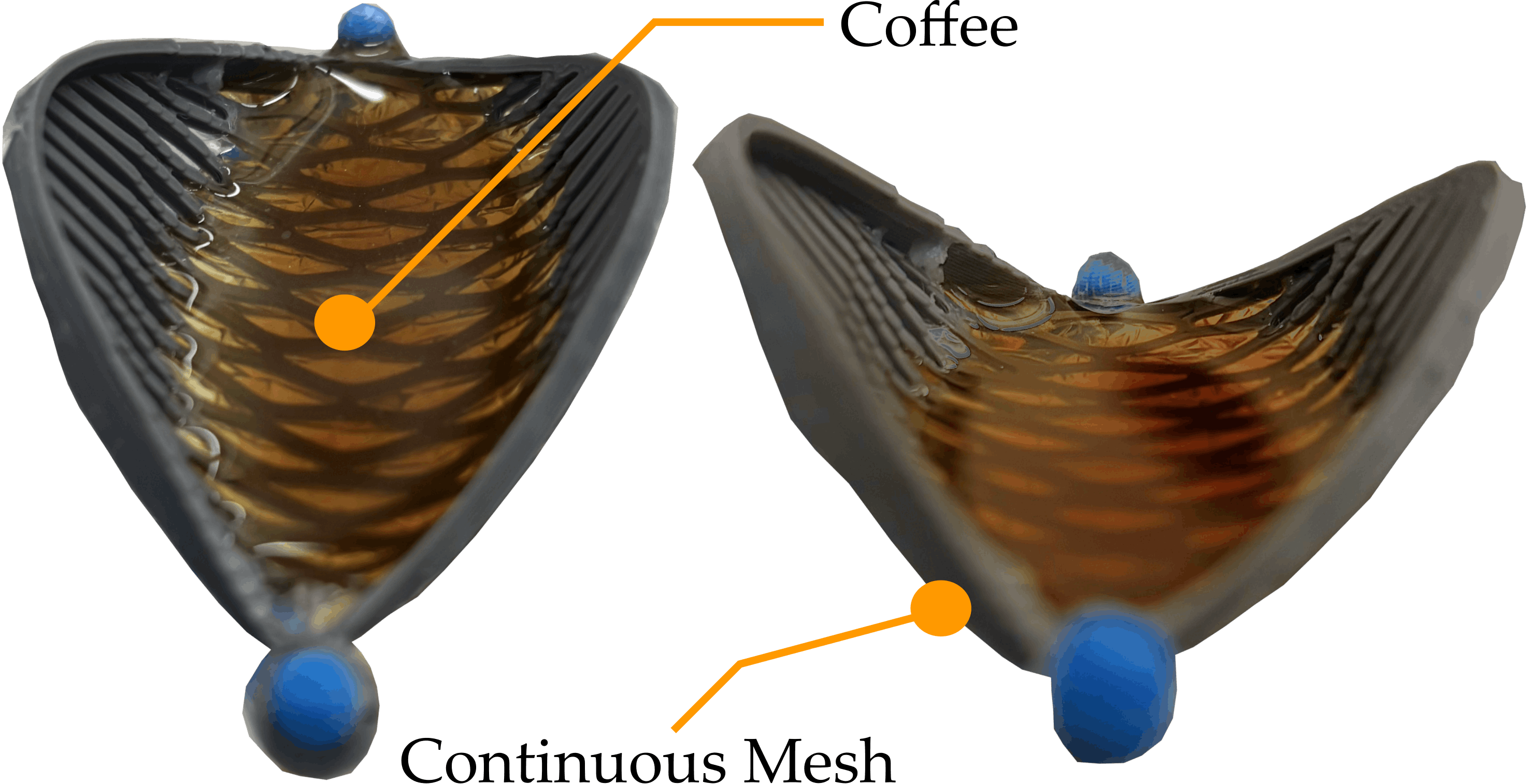}
		\caption{A prototype of the non-porous kiri-spoon. We $3$D-print the kirigami mesh on top of a plastic sheet to allow deformation while keeping the gaps sealed. This continuous mesh is necessary for liquid food such as soup.}
		\label{fig:mesh}
	\end{center}
 	\vspace{-1.5em}
\end{figure}

In this paper we re-designed the kiri-spoon using a pivot-based approach, enabling hand-held and robot-mounted versions.
These kiri-spoons act like pliers: by squeezing the handle, the user extends the kirigami mesh and encloses food morsels.
The fundamental advantage of our design is its \textit{accessibility}: users can $3$D print and snap together their own hand-held kiri-spoons with just four parts made of PLA and TPU.
By adding a servo motor, this same system can be mounted to a robot arm.
We collaborated with stakeholders throughout the design process.
In particular, stakeholders highlighted that closing the kiri-spoon was easy, but opening the kiri-spoon was difficult.
We therefore added a resistive band, and characterized the force-displacement relationship as a function of the kiri-spoon geometry and kirigami materials.
Designers can use our resulting models to personalize their own kiri-spoons to the needs of individual stakeholders.
Our preliminary experiments with users without disabilities suggested that the kiri-spoon improves robot-assisted feeding, and initial surveys with stakeholders with Parkinson's indicated that the hand-held kiri-spoon was comfortable, intuitive, and effective at preventing spills.

\p{Limitations}
Additional testing with a larger set of stakeholders is necessary to better understand the user's perspective, particularly regarding the deployment and safety of the robot-mounted kiri-spoon in residential care settings.
A core limitation of kirigami is that the ribbons make the mesh porous; when the kiri-spoon is actuated, gaps in the kirigami grow in size, allowing small food items to slip through.
This is problematic if users want to consume liquids like soup.
We show in \fig{mesh} that it is possible to address this issue by making a continuous mesh.
However, the food safety of the plastics used in this mesh is not well established.

%% file: bibtex.bib
@article{keely2025kiri,
  title={Kiri-{S}poon: {A} kirigami utensil for robot-assisted feeding},
  author={Keely, Maya and Franco, Brandon and Grothoff, Casey and Jenamani, Rajat Kumar and Bhattacharjee, Tapomayukh and Losey, Dylan P and Nemlekar, Heramb},
  journal={The International Journal of Robotics Research},
  pages={02783649251405291},
  year={2025}
}

@inproceedings{keely2024kiri,
  title={{Kiri-Spoon: A} soft shape-changing utensil for robot-assisted feeding},
  author={Keely, Maya N and Nemlekar, Heramb and Losey, Dylan P},
  booktitle={International Conference on Intelligent Robots and Systems},
  year={2024}
}

@article{liu2025robot,
  title={Robot-assisted feeding: {A} systematic review and future prospects},
  author={Liu, Fei and Li, Zhi and Hu, Mingyue},
  journal={Technology and Health Care},
  year={2025},
}

@article{park2020active,
  title={Active robot-assisted feeding with a general-purpose mobile manipulator: {D}esign, evaluation, and lessons learned},
  author={Park, Daehyung and Hoshi, Yuuna and Mahajan, Harshal P and Kim, Ho Keun and Erickson, Zackory and Rogers, Wendy A and Kemp, Charles C},
  journal={Robotics and Autonomous Systems},
  year={2020}
}

@inproceedings{nanavati2025lessons,
  title={Lessons Learned from Designing and Evaluating a Robot-Assisted Feeding System for Out-of-Lab Use},
  author={Nanavati, Amal and Gordon, Ethan K and Faulkner, Taylor A Kessler others},
  booktitle={International Conference on Human-Robot Interaction},
  year={2025}
}

@article{ETstats,
  title={Essential tremor},
  author={Welton, Thomas and Cardoso, Francisco and Carr, Jonathan A and Chan, Ling-Ling and Deuschl, G{\"u}nther and Jankovic, Joseph and Tan, Eng-King},
  journal={Nature Reviews Disease Primers},
  pages={83},
  year={2021}
}

@inproceedings{chang2025design,
  title={Design of a Breakaway Utensil Attachment for Enhanced Safety in Robot-Assisted Feeding},
  author={Chang, Hau Wen and Yow, J-Anne and Lim, Lek Syn and Ang, Wei Tech},
  booktitle={International Conference On Rehabilitation Robotics},
  year={2025}
}

@inproceedings{song2025soda,
  title={{SODA--S}oft Origami Dynamic Utensil for Assisted Feeding},
  author={Song, Yuxin Ray and Luo, Yiyue},
  booktitle={International Conference on Soft Robotics},
  year={2025}
}

@inproceedings{sundaresan2023learning,
  title={Learning Sequential Acquisition Policies for Robot-Assisted Feeding},
  author={Sundaresan, Priya and Wu, Jiajun and Sadigh, Dorsa},
  booktitle={Conference on Robot Learning},
  year={2023}
}

@article{sabari2019adapted,
  title={Adapted feeding utensils for people with {P}arkinson’s-related or essential tremor},
  author={Sabari, Joyce and Stefanov, Dimitre G and Chan, Judy and Goed, Lorraine and Starr, Joyce},
  journal={The American Journal of Occupational Therapy},
  year={2019}
}

@article{pagnussat2025people,
  title={How do people with {P}arkinson’s disease perceive challenges in handling cutlery?--{A} mixed study},
  author={Pagnussat, Aline Souza and Pinho, Alexandre Severo do and Pinto, Camila and Rosa, Thainara Cruz da and Moscovich, Mariana and de Sousa Andrade, Carine and Chen, Yi-An},
  journal={Disability and Rehabilitation: Assistive Technology},
  year={2025}
}

@manual{obi,
	title = "Obi: The Adaptive Eating Device",
    author = "Obi",
	note = "\url{https://meetobi.com/}",
	year = "2026"
}

@manual{neater,
	title = "Neater Eater Robotic",
    author = "Neater",
	note = "\url{https://www.neater.co.uk/neater-eater-robotic}",
	year = "2026"
}

@manual{rehabmarket,
	title = "Adaptive Eating Utensils",
    author = "RehabMarket",
	note = "\url{https://www.rehabmart.com/category/eating_utensils_and_accessories.htm}",
	year = "2026"
}

@manual{liftware,
	title = "Liftware Stabilizing Spoon",
    author = "{The Michael J. Fox Foundation for Parkinson's Research}",
	note = "\url{https://www.michaeljfox.org/news/apply-free-liftware-stabilizing-spoon}",
	year = "2026"
}

@manual{eli,
	title = "Independent Eating",
    author = "ELISpoon",
	note = "\url{https://elispoon.com/}",
	year = "2026"
}

@misc{VirginiaHomePurpose,
	title = "A Rich Tradition of Compassion",
    author = "{The Virginia Home}",
	note = "\url{https://thevirginiahome.org/}",
	year = "2026"
}

@misc{PLAinfo_ProtolabsNetwork,
    title = "What is {PLA} (polylactic acid) filament?",
    author = "{Protolabs Network}",
	note = "\url{https://www.hubs.com/knowledge-base/what-is-pla/}",
	year = "2026"
}

@misc{TPUinfo_ProtolabsNetwork,
    title = "What is {TPU} and when should you use it?",
    author = "{Protolabs Network}",
	note = "\url{https://www.hubs.com/knowledge-base/what-is-tpu-and-when-should-you-use-it/}",
	year = "2026"
}

@inproceedings{bhattacharjee2020more,
  title={Is more autonomy always better? {E}xploring preferences of users with mobility impairments in robot-assisted feeding},
  author={Bhattacharjee, Tapomayukh and Gordon, Ethan K and Scalise, Rosario and Cabrera, Maria E and Caspi, Anat and Cakmak, Maya and Srinivasa, Siddhartha S},
  booktitle={International Conference on Human-Robot Interaction},
  year={2020}
}

@inproceedings{jenamani2024flair,
  title={{FLAIR: F}eeding via Long-horizon AcquIsition of Realistic dishes},
  author={Jenamani, Rajat Kumar and Sundaresan, Priya and Sakr, Maram and Bhattacharjee, Tapomayukh and Sadigh, Dorsa},
  year={2024},
  booktitle={Robotics: Science and Systems}
}

@article{tai2025grits,
  title={{GRITS: A} Spillage-Aware Guided Diffusion Policy for Robot Food Scooping Tasks},
  author={Tai, Yen-Ling and Yang, Yi-Ru and Yu, Kuan-Ting and Chao, Yu-Wei and Chen, Yi-Ting},
  journal={arXiv preprint arXiv:2510.00573},
  year={2025}
}

@article{padmanabha2025waffle,
  title={{WAFFLE: A} Wearable Approach to Bite Timing Estimation in Robot-Assisted Feeding},
  author={Padmanabha, Akhil and Yuan, Jessie and Mehta, Tanisha and Jenamani, Rajat Kumar and Hu, Eric and de Le{\'o}n, Victoria and Wertz, Anthony and Gupta, Janavi and Dodson, Ben and Yan, Yunting and others},
  journal={arXiv preprint arXiv:2510.03910},
  year={2025}
}

@inproceedings{gordon2023towards,
  title={Towards general single-utensil food acquisition with human-informed actions},
  author={Gordon, Ethan Kroll and Nanavati, Amal and Challa, Ramya and Zhu, Bernie Hao and Faulkner, Taylor Annette Kessler and Srinivasa, Siddhartha},
  booktitle={Conference on Robot Learning},
  year={2023}
}

@article{tsakona2025continuous,
  title={Continuous Real-time Adaptation Framework for Enhancing Trust and Technology Acceptance: {A}n Assistive Feeding Study},
  author={Tsakona, Dimitra and Demiris, Yiannis},
  journal={Transactions on Human-Robot Interaction},
  pages={1--37},
  year={2025}
}

@article{martin2025anisotropic,
  title={Anisotropic and hyperelastic mechanical response of {3D} printed {TPU} parts},
  author={Mart{\'\i}n-Sosa, Ezequiel and T{\'a}vara, Luis and Ojeda, Joaqu{\'\i}n and Estefani, Alejandro},
  journal={Progress in Additive Manufacturing},
  pages={5697--5709},
  year={2025}
}

@inproceedings{yow2025franc,
  title={{FRANC: F}eeding Robot for Adaptive Needs and Personalized Care},
  author={Yow, J-Anne and Toh, Luke Thien Luk and San, Yi Heng and Ang, Wei Tech},
  booktitle={International Conference on Intelligent Robots and Systems},
  year={2025}
}
